# An improved LogNNet classifier for IoT applications

**H Heidari[1] and A A Velichko[1,3]**

[1] Department of Applied Mathematics, Damghan University, Damghan, Iran
[2] Institute of Physics and Technology, Petrozavodsk State University, 33 Lenin str., 185910, Petrozavodsk, Russia

[3] E-mail: velichko@petrsu.ru

**Abstract.** In the age of neural networks and Internet of Things (IoT), the search for new neural network architectures capable of operating on devices with limited computing power and small memory size is becoming an urgent agenda. Designing suitable algorithms for IoT applications is an important task. The paper proposes a feed forward LogNNet neural network, which uses a semi-linear Henon type discrete chaotic map to classify MNIST-10 dataset. The model is composed of reservoir part and trainable classifier. The aim of the reservoir part is transforming the inputs to maximize the classification accuracy using a special matrix filing method and a time series generated by the chaotic map. The parameters of the chaotic map are optimized using particle swarm optimization with random immigrants. As a result, the proposed LogNNet/Henon classifier has higher accuracy and the same RAM usage, compared to the original version of LogNNet, and offers promising opportunities for implementation in IoT devices. In addition, a direct relation between the value of entropy and accuracy of the classification is demonstrated.

## 1. Introduction

The fast development of smartphones, tablet PCs and wearable devices lead to an exponential increase in the volume of transferred images through networks. Image classification is an important tool for extracting information from digital images. This method is widely used in science and engineering. Murugappan proposed the method for facial emotional recognition [1]. Zhang et al introduced the synergic deep learning to classify medical diagnosis images [2]. Peng et al conducted the security assessment of deep learning for image classification [3].

The fourth industrial revolution (Industry 4.0) is the automation and data exchange in the official and industrial processes to improve various aspects of our lives. Industry 4.0 enables organizations and industries to digitize their systems, improve their performance and help its customers make better decisions regarding waste management and environmental sustainability. The internet of things is one of the building blocks of Industry 4.0. IoT focuses on the leveraging interconnection between the physical objects and users, which collect and share data through the internet. In recent years, IoT received wide attention from the scientists and manufactures due to its wide applications in industries. For example, IoT is used in health care, fitness, education, entertainment, social life, energy conservation, environment monitoring, home automation and transport systems [4]. In 2020, the global IoT industry market was 119 Billion dollars. The annual Compound Annual Growth Rate of 20% is predicted for IoT market [5].

In recent years, the IoT image classification attracts a great attention due to the huge number of IoT users and wide applications of image classification. However, the small memory size and high computational efforts are two important problems in the designing of IoT devices. Sun et al introduced the 22nm device of STT MRAM with processing-in-memory CNN accelerator [6]. Sohoni et al investigated the required memory for training a neural network [7]. Utrilla et al designed the resource-constrained IoT device based on gated recurrent unit neural network [8]. Bhardwaj introduced network of neural networks to overcome the low-memory bottleneck [9]. Penkovsky et al used the hybrid CMOS - hafnium oxide resistive memory technology to reduce the memory requirement [10]. Velichko designed LogNNet model to reduce the total used memory in an IoT device. LogNNet is feed forward neural network with the reservoir property which uses chaotic logistic map to classify MNIST-10 dataset [11]. This paper proposes modifications to Velichko study [11] to improve the accuracy, while the required memory is approximately unchanged.

The rest of the paper is organized as follows. In Section 2, the preliminaries are given. The proposed method is discussed in Section 3. The numerical results are given in Section 4. Section 5 is devoted to the conclusions and future studies.

## 2. Preliminaries

### 2.1. The concept of RAM saving on the LogNNet classifier

A new neural network architecture, named LogNNet, was proposed in [11], which allows efficient use of small amounts of RAM and opens up a new potential for designing low-power IoT devices. LogNNet is a feed forward neural network, where signals are directed only from input to output. The main feature of this network is using deterministic chaotic filters, which mix the input information in incoming signals. The network uses the discrete logistic map to generate a chaotic sequence and mixed input to extract initially invisible information.

The model saves RAM because the array of weights is not stored in the chaotic filter, and only three parameters are required to calculate the weights. LogNNet was tested on the MNIST-10 database and demonstrated the recognition accuracy of about 96.3%. The developed architecture uses no more than 29 kB of RAM. In addition, LogNNet was simulated on devices with very small RAM and it took only 1-2 kB of RAM. The results showed good accuracy of 80.3%. A miniature controller Atmega328, which fits on a fingertip and can be used in "smart" devices of small dimensions, has about the same amount of memory. Therefore, the developed network with a digital reservoir is a promising network architecture in comparison with the known algorithms. This paper proposes a feed forward LogNNet neural network, which uses a semi-linear Henon type discrete chaotic map to classify MNIST-10 dataset. The parameters of the chaotic map are optimized using particle swarm optimization (PSO) with random immigrants. The proposed LogNNet/Henon classifier has higher accuracy and same RAM usage, compared to the original version of LogNNet, and has broad opportunities for implementation in IoT devices.

### 2.2. The concept of chaos on the LogNNet classifier

Chaotic systems are deterministic nonlinear dynamical systems, which behavior is sensitive to the initial conditions. A little change in the initial conditions may lead to big difference in their trajectories. In recent years, chaotic systems attracted wide attention from theoretical and practical point of view. Fan et al studied long-term prediction of chaotic systems with machine learning [12]. Velichko applied chaotic logistic map for handwriting digits classification [11]. Luo et al introduced a four dimensional chaotic system and used it for image encryption [13].

The degree of chaos in the reservoir part of LogNNet affects the efficiency of the model [11]. The classification accuracy depends on the parameter of logistic maps. The form of the dependence correlates with the dependence of the Lyapunov exponent $\lambda$ on the parameter r, and indicates that the chaotic dynamics of the logistic map plays an important role in the recognition process by the neural network.

Lyapunov exponent is a parameter for investigating the stability of dynamical system. Entropy is another important concept that measures the rate of the complexity in dynamical system [14]. In this study, we show that entropy is a better indicator correlating with classification accuracy of LogNNet network.

## 3. Methods

### 3.1. Modification of LogNNet classifier with Henon type chaotic map

Researchers believe that designing a neural network is an art. This paper presents an optimized feedforward LogNNet neural network. The model consists two parts: reservoir part and trainable classifier. A schematic representation of the LogNNet architecture is shown in **Ошибка! Источник ссылки не найден.**. An image of a handwritten digit from the MNIST-10 database is the network input. Based on [11], the pattern T3 is selected to convert the 28×28 pixel image into a one-dimensional array $Y[i]$, where $i = 0 - 784$, and $Y[0]=1$.

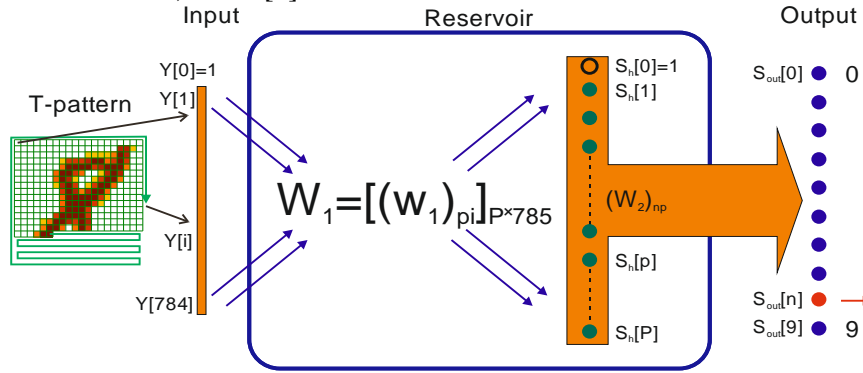

**Figure 1**. The model structure.

The first part is the reservoir of the model. It transfers the input array $Y$ into a more suitable vector $S_h$ which is used in the training process of the neural network. The function of the reservoir part, denoted by $f_h$ is following

$$S_h = f_h(W_1 \cdot Y),\qquad(1)$$

where $W_1$ is the given matrix of input weights. The dimension of the matrix $W_1$ is $P \times 785$, where $P$ is the number of neurons in the reservoir.

The reservoir part has two aims. The first aim is to transform the vector $Y$ into vector $S_h$ with the dimension $P$ in a way that the best classification accuracy is achieved. The second aim is to reduce the required RAM, which is critical in IoT applications [11].

The training process and the final output are done in the second part of the model, denoted by $S_{out}$, as follows

$$S_{out} = f_{out}(W_2 \cdot S_h).\qquad(2)$$

The output layer $S_{out}$ contains $N = 10$ neurons and determines the result of image classification from the MNIST-10 database. The $W_2$ matrix is trained by the back propagation method. If the output classifier contains one layer of neurons, then the layer architecture of the neural network is denoted as LogNNet 784:$P$:$N$. If the output classifier contains two layers with an additional hidden layer with the number of neurons $H$, then the architecture of the neural network is denoted as LogNNet 784:$P$:$H$:$N$. In the latter case, an additional matrix of weights ($W_3$) is required.

We construct the weight matrix $W_1$ using the following discrete chaotic map [15]

$$\begin{cases} x_{n+1} = y_n \\ y_{n+1} = x_n + a_1 x_n^2 + a_2 y_n^2 - a_3 x_n y_n - a_4 \end{cases},\qquad(3)$$

where $\{a_i,\ i=1,\dots,4\}$ are given constants, $n \in \mathbb{N}$ and $x_0,\ y_0$ are the given initial conditions. This function is a variation of Henon map, which has hidden chaotic attractor for some values of $\{a_i,\ i=1,\dots,4\}$. The matrix $W_1$ is constructed using $y_{n+1}$ for a predefined number $n$. Our experiments show that the accuracy of the classification is very sensitive to the element values of the matrix $W_I$ and to the method of matrix filling (see subsection 4.1). Therefore, specifying the values of {$a_i$, $i$=1,…,4} is an important part in designing the neural network.

### 3.2. Particle swarm optimization
Particle swarm optimization is a metaheuristics optimization technique, proposed by Kennedy and Eberhart [16]. This algorithm is inspired by swarm behavior, such as bird flocking and schooling fish in food searching. The swarm members are sharing information to achieve the best result. From the mathematical point of view, the algorithm is simple and easy to implement. Therefore, it is suitable to use in the IoT devices with memory limitations. In PSO, each particle $i$ is characterized by its location $x_i$ and the velocity $v_i$. To improve the current solution, the location and velocity of the particles are changed in each iteration in the following way

$$x_{i+1} = x_i + v_i \tag{4}$$

$$v_{i+1} = \omega v_i + c_1 r_1 (x_{Best_i} - x_i) + c_2 r_2 (g_{Best} - x_i), \tag{5}$$

where $x_{Best}$ and $g_{Best}$ denote the best position of particle $i$ and the global best position of all particles until the current iteration, respectively. The parameters $r_1$ and $r_2$ are random numbers between 0 and 1. The parameter $\omega$ is inertia weight and the parameters $c_1$ and $c_2$ are individual-cognition parameters, which determine the effects of the particle's own previous experiences on the global optimal solution [16,17]. These parameters play an important role in the convergence of PSO algorithm. Our experiments demonstrate that $c_1 = c_2 = 2$ and $\omega = 0.5$ are suitable values for the proposed problem. Recently, Ünal and Kayakutlu used random immigration method to improve the accuracy of PSO in multi-objective optimization problem [17]. In this paper, we use a similar method, where 70% of particles are chosen randomly in each iteration. This method is denoted as RPSO.

### 3.3. Optimization of parameters
Training, validation and test are the main processes in every machine-learning algorithm.

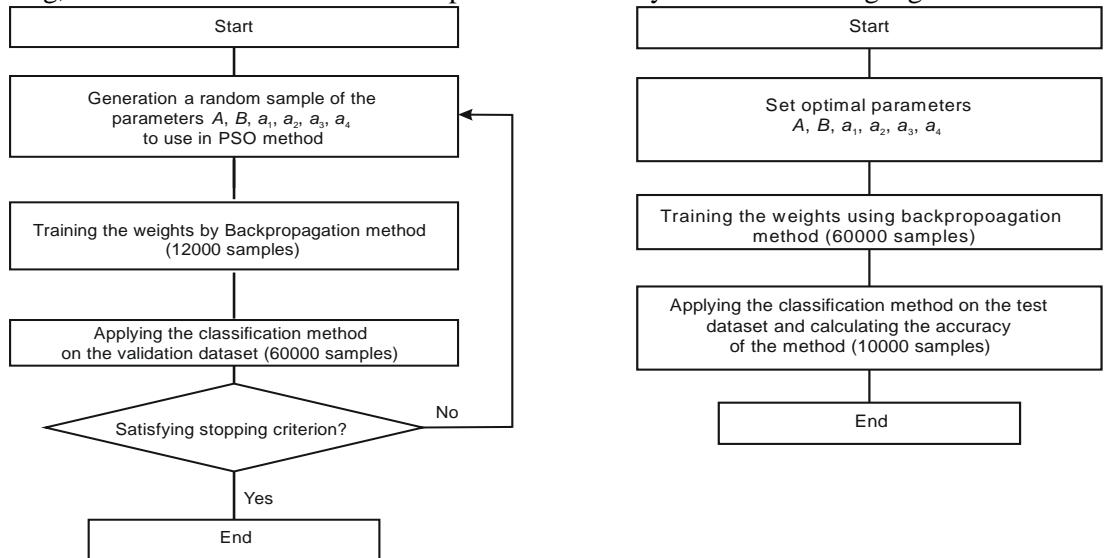

**Figure 2**. Parameters optimization algorithm in the reservoir part of the proposed model.

**Figure 3**. The test process of the proposed model.

The aim of validation process is fine-tuning the performance of the training process in neural network. The test process determines the accuracy and margin of error of the neural network. This paper uses a two-stage algorithm to construct LogNNet model. In the first stage, the PSO method is applied to find the optimized coefficients of the Hennon type map. In the second stage, the MNIST-10 dataset and the results of the first stage are used to train the neural network.

The matrix $W_1$ is constructed in the reservoir part to transfer the input of the model in a way that the accuracy is maximized. The elements of the matrix $W_1$ depend on the values of $\{a_i, i=1,…,4\}$ and the initial conditions $(x_0, y_0)$ in the Henon type map (Eq. (3)). We use RPSO method for determining the best values of these parameters to maximize the accuracy of the model. To speed up the training process during optimizing the parameters in the reservoir part (the first stage), twenty percent of the training dataset is considered (12000 objects) (see **Ошибка! Источник ссылки не найден.**). All elements of the training dataset (60000 objects) are used in validation process. After optimization, a re-training process was carried out on a full base (60,000 objects), followed by final testing using a test base of 10,000 objects (see figure 3). As a result, the classification accuracy was optimal for the given configuration of the LogNNet / Henon network.

## 4. Numerical results

### 4.1. Classification and matrix filling methods
Based on Section 3, the initial conditions $(x_0, y_0)$ are needed to construct the weight matrix $W_1$. We consider two forms of the initial conditions. The first form is initial condition of the sine function as in [11]. The first row of the matrix $W_1$ is constructed in the following way

$$\begin{cases} W_1[1,i] = x_0 = A \sin(\dfrac{i}{784} \cdot \dfrac{\pi}{B}), \\ y_0 = 0.51 \end{cases}$$

(6)

where $A$ and $B$ are given constant numbers. The next rows are constructed by using the previous rows as the input of Eq. (3). The second form is the constant initial conditions $(x_0, y_0)=(A, B)$, and the matrix $W_1$ is filled in a snake manner $(W_1[p,i]=y_n)$, as depicted in figure 4(a). Its pseudo code is presented in figure 4(b).

To avoid the influence of transient period at constant initial conditions [18], the preliminary 10000 iterations are required.

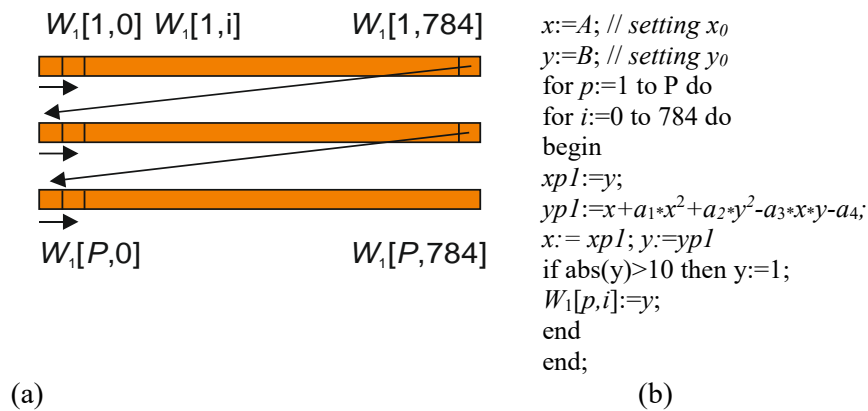

(a)                                                                                           (b)

**Figure 4**. The method for constructing the matrix weights for the constant initial conditions (a) Graphical representation, (b) The pseudo code of the applied algorithm ($x$ and $y$ represent $x_n$ and $y_n$, $xp1$ and $yp1$ represent $x_{n+1}$ and $y_{n+1}$ in (3)).

In addition, the following upper limit condition for $y_n$ is applied to investigate its effect on the accuracy of the classification

$$If \ |y_n| > 10 \ then \ y_n = 1.$$

(7)

The hypothesis (7) is required, as the time series may tend to an unstable mode of abnormally large numbers sequence, generated by certain values of the parameters ($A$, $B$, $a_1$, $a_2$, $a_3$, $a_4$).

The methods numbered from 1 to 6 are explained in Table 1.

**Table 1.** The properties of the different methods.

| Method<br>Property | 1 | 2 | 3 | 4 | 5 | 6 |
|---|---|---|---|---|---|---|
| Initial conditions | Sine function | Sine function | Constant numbers | Constant numbers | Constant numbers | Constant numbers |
| Preliminary 10000 iterations | | | Yes | Yes | No | No |
| Upper limit condition (Eq. (7)) satisfied. | No | Yes | Yes | No | Yes | No |

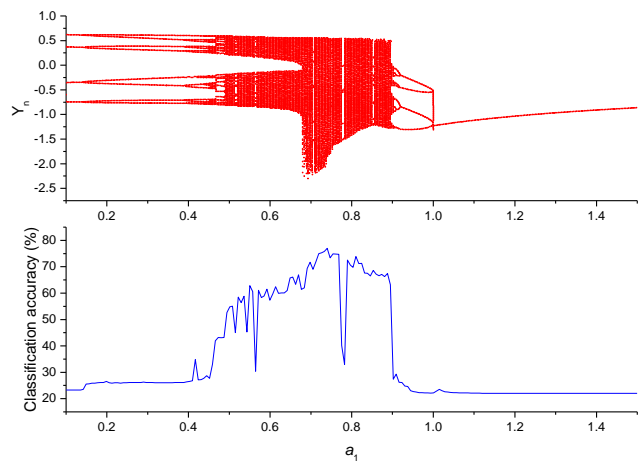

**Figure 5**. The bifurcation diagram and classification accuracy for different values of $a_1$, when $A$=-0.81, $B$=0.51, $a_2$=1, $a_3$=1.51, $a_4$=0.74 and the Method 6 is used for LogNNet/Henon 784:25:10.

The figure 5 shows the bifurcation diagram of the Henon type map and the accuracy of LogNNet (method 6) for different values of $a_1$. The variation of $a_1$ affects the classification accuracy over a wide range, while the other parameters are fixed. To find the parameters with the highest classification accuracy, we used the RPSO method.

A detailed comparison between LogNNet and other classification methods is done in [11], and here, we compare the proposed method with LogNNet. The activation function and the layers architecture of the feedforward neural model of the proposed method are selected the same as in [11]. The results for the different proposed models are given in the Table . RPSO runs with 150 particles in 100 iterations, and the lower and upper bounds for the variables [$a_1$,$a_2$,$a_3$,$a_4$] are set [0.01,0.1,0,0,0,0] and [1.5,10,1.5,1.5,1.5,1.5], respectively. The neural network model is trained with the maximum 20 epochs in optimization mode (see. figure 2).

The differences between the accuracy of the proposed methods and [11] are given in table 3. The proposed methods have significantly better accuracy than [11]. From a practical point of view, we suggest to use LogNNet/Henon Method 4 or LogNNet/Henon Method 6, as they have better performance in all considered layer structures.

**Table 2.** The accuracy of classification for different proposed methods.

| Layer architecture | LogNNet [11] | LogNNet /Henon Method 1 | LogNNet /Henon Method 2 | LogNNet /Henon Method 3 | LogNNet /Henon Method 4 | LogNNet /Henon Method 5 | LogNNet /Henon Method 6 |
|---|---|---|---|---|---|---|---|
| 784:25:10 | 80.3% | 84.02% | 83.75% | 84.54% | 84.7% | 83.24% | 84.68% |
| 784:100:10 | 89.5% | 89.90% | 90.13% | 90.8% | 90.78% | 91.02% | 90.65% |
| 784:200:10 | 91.3% | 91.21% | 90.94% | 90.8% | 92.00% | 91.02% | 91.78% |
| 784:100:60:10 | 96.3% | 96.21% | 96.51% | 96.78% | 96.84% | 96.67% | 97.09% |

**Table 3.** The difference between the proposed methods and [11].

| Layer architecture | LogNNet/ Henon Method 1 | LogNNet/ Henon Method 2 | LogNNet/ Henon Method 3 | LogNNet/ Henon Method 4 | LogNNet/ Henon Method 5 | LogNNet/ Henon Method 6 |
|---|---|---|---|---|---|---|
| 784:25:10 | 3.72% | 3.45% | 4.24% | 4.4% | 2.94% | 4,38% |
| 784:100:10 | 0.4% | 0.63% | 1.3% | 1.28% | 1.52% | 1.15% |
| 784:200:10 | -0.09% | -0.36% | -0.50% | 0.7% | -0.28% | 0.48% |
| 784:100:60:10 | -0.09% | 0.21% | 0.48% | 0.54% | 0.37% | 1.06% |

*4.2. Relation between the entropy and accuracy of the classification*

Another interesting result is the direct relation between the approximate entropy (ApEn) [19] of the time series filling matrix $W_1$ and accuracy of the classification. ApEn has special parameters $m$ and $r$. Let us use the method 4 with $(x_0, y_0)$=(-0.81,0.51), $a_2$=1, $a_3$=1.51, $a_4$=0.74 and the parameter $a_1$ set between 0.1 and 1.5. The classification accuracy for different values of $a_1$ is given in **Ошибка! Источник ссылки не найден.**. The same figure shows the relation of ApEn for different values of m and r for a time series. Let us consider the points $p_1$, $p_2$, $p_3$, $p_4$, $p_5$, which are depicted in this figure. A comparison between the subfigures of approximate entropy shows that the subfigure with $r$=0.025 and $m$=2 match with the trend of classification accuracy, except the point $p_5$, where the maximum accuracy is achieved. The reason is that ApEn is sensitive to the input parameters ($m$ and $r$) and is accurate in measuring the complexity of time series [20]. The Poincare plots for the points $p_3$, $p_4$ and $p_5$ are depicted in figure 7. The complexity of the time series is highest in the point $p_5$, where the classification accuracy is highest too. Therefore, applying the LogNNet method on MNIST-10 dataset is as an alternative way to measure the regularity of the time series, and it provides more reliable results than the ApEn calculation method.

## 5. Conclusion and future studies

The paper proposes a new structure for the reservoir part of feedforward neural network LogNNet. Results demonstrate that the accuracy of the model depends on the initial conditions and the chosen discrete chaotic map used to construct the weight matrix $W_1$. A Henon type chaotic map is examined, and RPSO method is used for determining the best parameters of the map to maximize the accuracy of the classification. The memory requirements for IoT devices remain the same. The proposed method uses the discrete chaotic map (3), which has four parameters and two parameters for initial conditions. The reference method [11] used the discrete chaotic logistic map, which required one parameter and two parameters for initial condition. The difference of required memory between the proposed method and [11] is the allocated memory for three real variables that is a minor difference. The accuracy of IoT devices can be improved by modifying the chaotic map and PSO method, while the training process and hardware used are unchanged. This method modification would help manufacturers to improve their products without additional hardware costs.

The results show that the accuracy is significantly improved in the layer structure 784:25:10. The layer structure 784:25:10 is the simplest structure, which requires small memory and is easy to implement in various IoT devices. The simple structure and low memory usage make the proposed LogNNet model suitable for implementation in edge computing in IoT devices, which will be a hot topic in near future [21]. The relation between the approximate entropy and the accuracy of the classification is investigated. Therefore, images classification by LogNNet can be considered as an alternative way to measure the regularity of the time series.

For the future studies, the different discrete chaotic maps with higher entropy can be examined to improve the accuracy. Applying the proposed method to different datasets for generalizing the results is another topic for the further research.

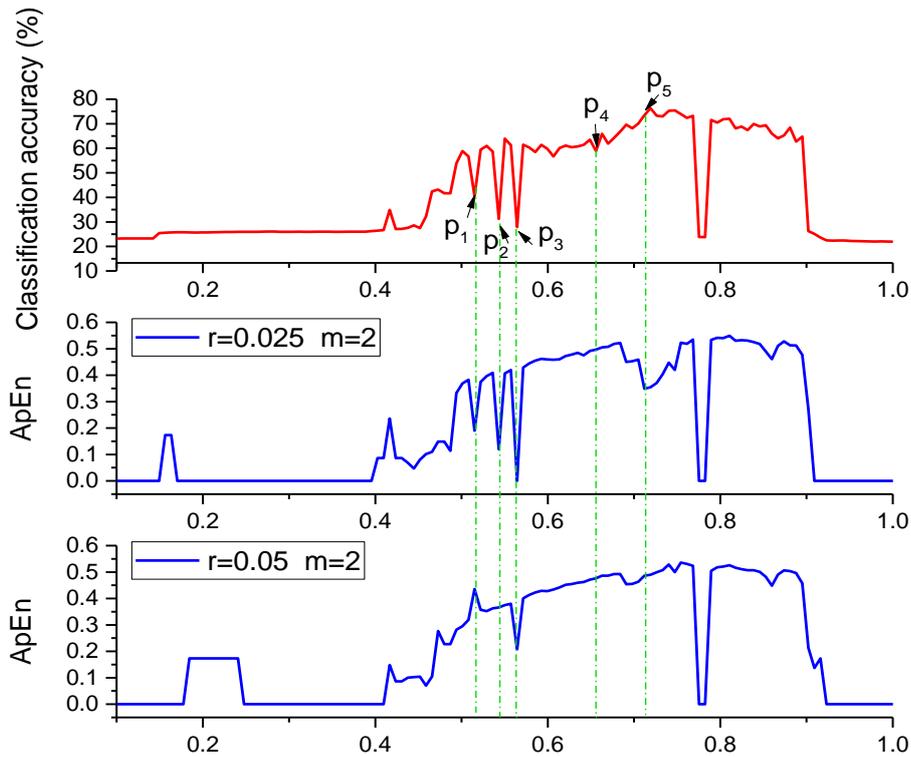

**Figure 6**. The classification accuracy and approximate entropy for different values of $a_1$ when $A$=-0.81, $B$=0.51, $a_2$=1, $a_3$=1.51, $a_4$=0.74 and the method 4 is used for LogNNet/Henon 784:25:10.

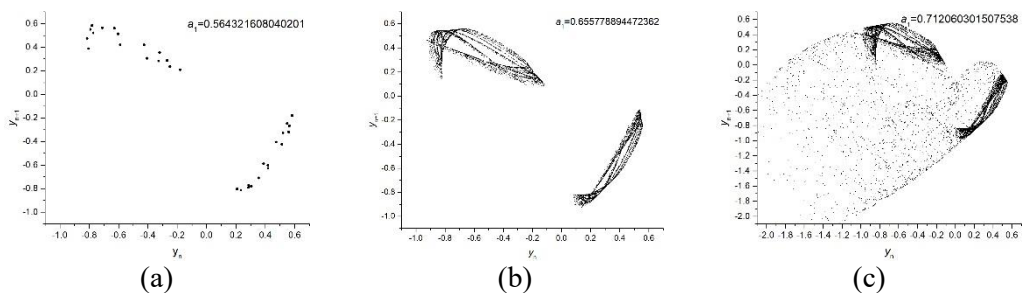

**Figure 7**. The Poincare plot for (a) the point $p_3$, (b) the point $p_4$, (c) the point $p_5$.

**Acknowledgments:** Authors express their gratitude to Andrei Rikkiev for the valuable comments in the course of the article's translation and revision.

## References

[1]   Murugappan M and Mutawa A 2021 Facial geometric feature extraction based emotional expression classification using machine learning algorithms *PLoS One* **16** e0247131

[2]   Zhang J, Xie Y, Wu Q and Xia Y 2019 Medical image classification using synergic deep learning *Med. Image Anal.* **54** 10–9

[3]   Peng Y, Member S, Zhao W, Cai W, Su J and Han B 2020 Evaluating Deep Learning for Image Classification in Adversarial *EICE Trans. Inf. Syst.* **103** 825–37

[4]   Sethi P and Sarangi S R 2017 Internet of Things : Architectures , Protocols , and Applications *J. Electr. Comput. Eng.* **2017** 9324035

[5]   Al-Sarawi S, Anbar M, Abdullah R and B. Al Hawari A 2020 Internet of Things Market Analysis Forecasts, 2020–2030 *2020 Fourth World Conference on Smart Trends in Systems, Security and Sustainability (WorldS4)* (London) pp 449–53

[6]   Sun B, Liu D, Yu L, Li J, Liu H, Zhang W and Torng T 2018 MRAM Co-designed Processing-in-Memory CNN Accelerator for Mobile and IoT Applications *arXiv* 1811.12179

[7]   Sohoni N S, Aberger C R, Leszczynski M, Zhang J and Ré C 2019 Low-memory neural network training: A technical report *arXiv* 1904.10631

[8]   Utrilla R, Fonseca E, Araujo A and Dasilva L A 2020 Gated Recurrent Unit Neural Networks for Automatic Modulation Classification with Resource-Constrained End-Devices *IEEE Access* **8** 112783–94

[9]   Bhardwaj K, Lin C, Sartor A and Marculescu R 2019 Memory- and communication-aware model compression for distributed deep learning inference on IoT *ACM Trans. Embed. Comput. Syst.* **18** 1–22

[10]  Penkovsky B, Bocquet M, Hirtzlin T, Klein J O, Nowak E, Vianello E, Portal J M and Querlioz D 2020 In-Memory Resistive RAM Implementation of Binarized Neural Networks for Medical Applications *Proceedings of the 2020 Design, Automation and Test in Europe Conference and Exhibition, DATE 2020* pp 690–5

[11]  Velichko A 2020 Neural Network for Low-Memory IoT Devices and MNIST Image Recognition Using Kernels Based on Logistic Map *Electronics* **9** 1432

[12]  Fan H, Jiang J, Zhang C, Wang X and Lai Y-C 2020 Long-term prediction of chaotic systems with machine learning *Phys. Rev. Res.* **2** 1–6

[13]  Luo Y, Zhou R, Liu J, Cao Y and Ding X 2018 A parallel image encryption algorithm based on the piecewise linear chaotic map and hyper-chaotic map *Nonlinear Dyn.* **93** 1165–81

[14]  Kapitaniak T, Mohammadi S A, Mekhilef S, Alsaadi F E, Hayat T and Pham V T 2018 A new chaotic system with stable equilibrium: Entropy analysis, parameter estimation, and circuit design *Entropy* **20** 670

[15]  Ouannas A, Wang X, Khennaoui A-A, Bendoukha S, Pham V-T and Alsaadi, E. F 2018 Fractional Form of a Chaotic Map without Fixed Points: Chaos, Entropy and Control *Entropy* **20** 720

[16]  Asma S and Heidari H 2018 Finite Time Mix Synchronization of Delay Fractional-Order Chaotic Systems *Glob. Anal. Discret. Math.* **3** 33–52

[17]  Ali Nadi Ünal and Kayakutlu G 2020 Multi-objective particle swarm optimization with random immigrants *Complex Intell. Syst. Vol.* **6** 635–50

[18]  Bao H, Wang N, Bao B, Chen M, Jin P and Wang G 2018 Initial condition-dependent dynamics and transient period in memristor-based hypogenetic jerk system with four line equilibria *Commun. Nonlinear Sci. Numer. Simul.* **57** 264–75

[19]  Pincus S M 1991 Approximate entropy as a measure of system complexity *Proc. Natl. Acad. Sci.* **88** 2297–301

[20]  Delgado-Bonal A and Marshak A 2019 Approximate entropy and sample entropy: A comprehensive tutorial *Entropy* **21** 541

[21]  Izotov Y A, Velichko A A, Ivshin A A and Novitskiy R E 2021 Recognition of handwritten MNIST digits on low-memory 2 Kb RAM Arduino board using LogNNet reservoir neural network *IOP Conf. Ser. Mater. Sci. Eng.* **1155** 12056